\documentclass[a4paper, 10pt, conference]{ieeeconf}      

\IEEEoverridecommandlockouts                              

\usepackage{makeidx}  
\usepackage{etoolbox} 
\usepackage{graphicx}
\usepackage{graphics}
\usepackage{amsmath}
\usepackage{url}
\usepackage[vlined, ruled, boxed, resetcount]{algorithm2e}
\SetKwProg{Function}{}{}{}
\usepackage{float}
\usepackage{hyperref}
\usepackage{stfloats}
\begin{document}
%
%

%
\title{\LARGE \bf 

A Robust and Rapidly Deployable Waypoint Navigation Architecture for Long-Duration Operations in GPS-Denied Environments


}

%
%
\author{Erik Pearson$^{1}$ \and Brendan Englot$^{1}$ \thanks{$^{1}$E. Pearson and B. Englot are with the Dept. of Mechanical Engineering, Stevens Institute of Technology, Hoboken, NJ 07030, USA, {\tt\small \{epearson, benglot\}@stevens.edu}. This work was supported by a grant from the Consolidated Edison Company of New York, Inc.} }
%
%
%

\maketitle              
\thispagestyle{empty}
\pagestyle{empty}

\begin{abstract}
For long-duration operations in GPS-denied environments, accurate and repeatable waypoint navigation is an essential capability.  
While simultaneous localization and mapping (SLAM) works well for single-session operations, repeated, multi-session operations require robots to navigate to the same spot(s) accurately and precisely each and every time. Localization and navigation errors can build up from one session to the next if they are not accounted for. Localization using a global reference map works well, but there are no publicly available packages for quickly building maps and navigating with them. We propose a new architecture using a combination of two publicly available packages with a newly released package to create a fully functional multi-session navigation system for ground vehicles. The system takes just a few hours from the beginning of the first manual scan to perform autonomous waypoint navigation. 
\end{abstract}
\section{Introduction}

\begin{figure}[ht]
    \centering
    \includegraphics[width=0.5\textwidth]{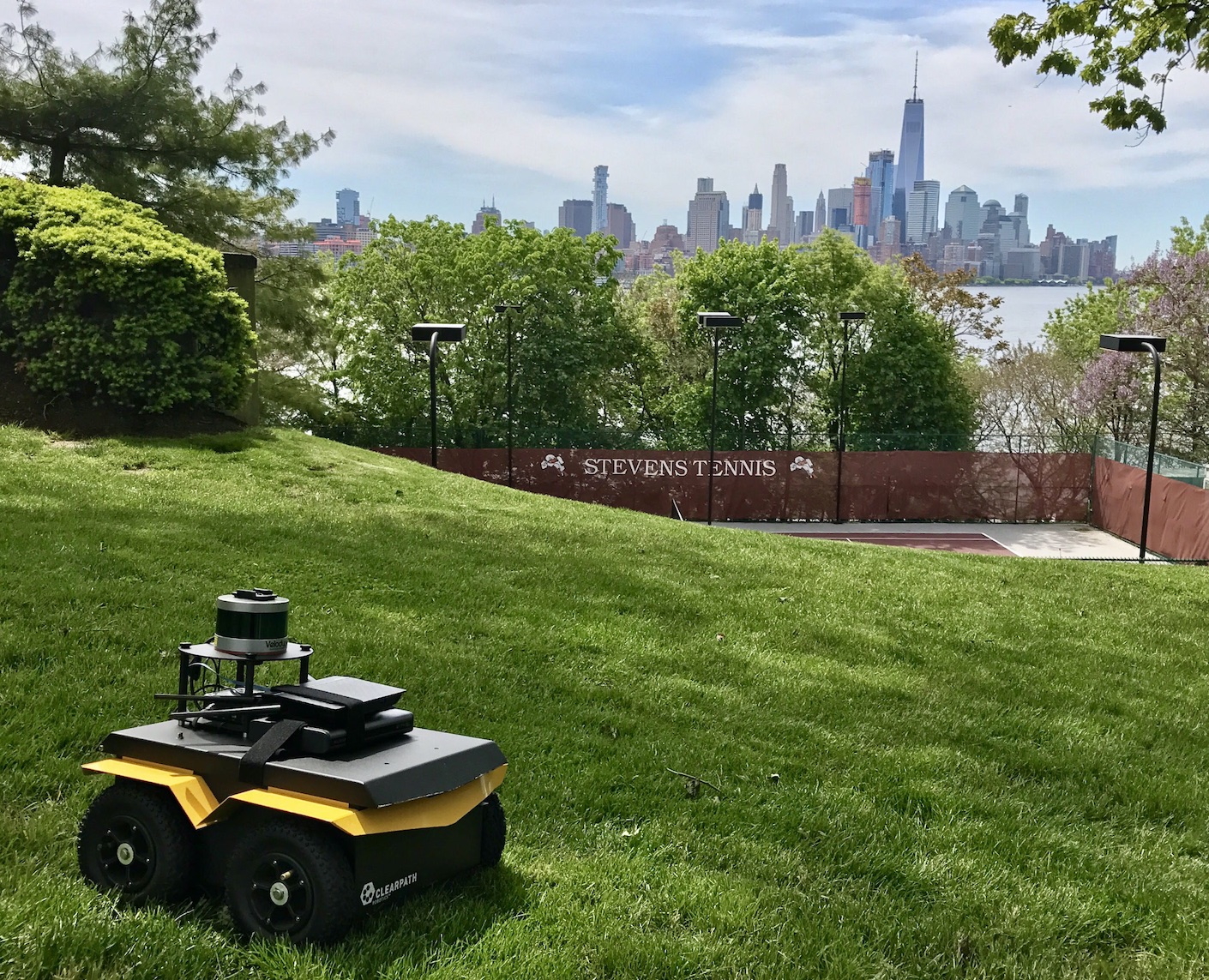}
    \caption{The Clearpath Jackal unmanned ground vehicle (UGV) used in this research, equipped with a Velodyne VLP-16 ``puck" lidar.}
    \label{fig:jackal}
\end{figure}

Mobile robots are often programmed for repeatable tasks, and each instance typically requires the same code. However, repeatable tasks require consistency between attempts, and localization is an important contributing factor to this consistency. For unmanned ground vehicles (UGVs) like Clearpath's Jackal seen in Fig. \ref{fig:jackal}, reliable navigation to specified waypoints can facilitate a wide range of repeatable tasks. For example, mobile manipulator platforms would be able to perform repeatable grasping and manipulation tasks at specified locations. 
However, there are currently no simple, publicly available localization methods and implementations compatible with repeated waypoint navigation that  incorporate fast map construction from scratch.

Simultaneous localization and mapping (SLAM) has become an essential tool for mobile robotics, and successful approaches allow vehicles to navigate around a previously unknown environment with confidence. However, many repeatable tasks occur in indoor environments that undergo few, if any, changes to their structure. Creating a new map on each attempt is unnecessary and time consuming for simple tasks. Additionally, new maps may not have the same orientation or origin as prior maps, resulting in different outcomes of repeated tasks. While key locations in a map can sometimes be identified semantically, the added complexity of doing so may sometimes be undesirable on embedded systems.

Localization using a prior map will ensure waypoints are placed at the same global location every time. Some common packages exist for SLAM within the Robot Operating System (ROS) \cite{quigley2009ros}. Currently there are no packages for localization using a prior map that also include navigation, therefore we have implemented a fast approach for building and using a prior map with a localization package for waypoint navigation in ROS, which is described in detail below.

\section{Related Work}
Mobile robot navigation has been researched extensively to open a pathway for more advanced tasks. However, navigation requires both localization and environmental data to make informed decisions. SLAM has been successful at satisfying both of those requirements. Most SLAM algorithms define a map origin based on the initial position of the robot when it begins its operations. To perform multi-session tasks, we need to ensure a common origin is used for each session.
\subsection{LiDAR-aided Pose Estimation}
Autonomous ground vehicles rely on accurate pose estimation for navigation \cite{weinstein2010pose, wooden2010autonomous}. While there are many methods for performing pose estimation \cite{chilian2011multisensor}, simplifying the task down to a common sensor type reduces the complexity. LiDAR sensors provide enough data for localization with comparative algorithms. Managing thousands of data points from each scan can be daunting from a computational complexity standpoint, however there are methods to reduce the complexity such as semantic labeling. This approach labels and groups a subset of point cloud data together to appear as a single entity, thereby significantly reducing the overall number of comparisons. Using semantics and Random Sample Consensus (RANSAC), \cite{wang2019robust} performs stable and accurate pose estimation while eliminating dynamic obstacles from comparisons. Semantic modeling can be a powerful tool for pose estimation as proved in \cite{ratz2020oneshot}, which solved global localization by registering a single LiDAR scan overlapped with a camera to a reference map using segmentation and neural network training. Localization can be performed by focusing on one semantic object class while attaining high accuracy in handling the surrounding data \cite{oelsch2022ro}.
\subsection{Multi-Session SLAM}
Two forms of multi-session SLAM persist in mobile robotics research. The first expands the boundaries of a previously defined map \cite{labbe2014online}, while the second revisits the same location from a previous map, where the environment may have changed \cite{mcdonald2013real}. Real world environments are not static, which requires maps to be updated from time to time for accurate localization. Therefore, there needs to be some tolerance for robots to use prior maps with outdated information, by either having enough static points of reference on a prior map or updating a prior map during each session. One example of enough reference data is the work produced by Labbe et al. \cite{labbe2022multi} in illumination invariant visual SLAM, where distinctly different visual references are capable of localizing successfully. Another method separates changes between the current map and the prior \cite{egger2018posemap}. Other researchers prefer to update the global reference map such as Zhao et al. \cite{zhao2021general}, who acknowledge active environment changes such as new stores within a mall should be manageable. 
\subsection{Repeated Navigation}
While consistent localization through pose estimation is required for multi-session waypoint navigation, environments are rarely static. One method to handle this is ignoring data unrelated to localization. Indoor artificial landmarks such as fiducial markers can be placed in an environment where other features may change \cite{dzodzo2013realtime}. Visual teach and repeat methods \cite{mattamala2022efficient, krajnik2018navigation} allow for repeated navigation without localization which can function even 
under seasonal environmental changes \cite{piasco2018survey, krajnik2017image}. However, 
robust localization methods can sometimes handle large static changes in the environment.

\subsection{Model Localization}
Prior maps can be built from many different data sources. One helpful source is 3D models constructed using computer aided design (CAD). Building Information Models can also be used to generate maps for both geometric and semantic localization \cite{yin2022semantic}. Not many real world structures have complete 3D models, however a 3D mesh can be approximately generated from a 2D floorplan for precise robot localization \cite{blum2020precise}. 3D models of real world environments work well in scenarios where the models already exist. For scenarios that begin with a completely unknown environment, the goal of this paper is to provide an easy-to-deploy alternative solution.
%

The main contributions of this work are:
\begin{itemize}
    \item Simple to use and quick to implement prior map localization with globally referenced waypoints,  for repeated mobile robot navigation tasks.
    \item An autonomous waypoint distributor package publishing 2D waypoint locations for repeatable navigation.
    \item A 3D localization package  robust enough to handle sources of error caused by differences in current LiDAR data and global reference maps, due to dynamic objects, displaced static objects and occlusions.
    \item A recommended package for building a global reference map quickly and accurately.
\end{itemize}
This unique framework will enable researchers to minimize time spent building custom solutions for repeated waypoint navigation tasks. We hope our implementation can be a stepping stone for work addressing highly complex tasks.

\section{Architecture Description}

To ensure accurate waypoint navigation, a global reference map and some method of performing localization are required. Global reference maps can be composed of labeled data, such as landmarks, or raw metric data, such as a point cloud. Many localization methods rely on a specific type of global reference map to perform optimally. SLAM algorithms build global reference maps from scratch.

Our framework uses publicly available packages to create a global reference map from a prior manual expedition, followed by localization for waypoint navigation in multi-session tasks. The package we selected for creating a global reference map was \textit{hdl\_graph\_slam}\footnote{https://github.com/koide3/hdl\_graph\_slam} \cite{koide2019portable} for two main reasons. The SLAM results were successful even when only LiDAR point cloud data was given, which reduced the requirements for data collection. There is also a service included that allows users to save a copy of the completed global map in the correct data type for the subsequent localization package.

The real-time localization package we chose is a publicly available Iterative Closest Point (ICP) \cite{besl} implementation from ETH Zurich\footnote{https://github.com/leggedrobotics/icp\_localization}. With some minor modifications for our specific systems, their package was able to take an initial position for the sensor used and accept odometry data for an estimate before performing ICP localization between an active sensor and the global reference map. With further testing, this localization method performed well in cases where the sensor had partial occlusion, dynamic obstacles moved around, and even when portions of the global reference map had been changed, such as moved boxes or chairs.

\begin{figure}[t]
    \centering
    \includegraphics[width=0.49\textwidth]{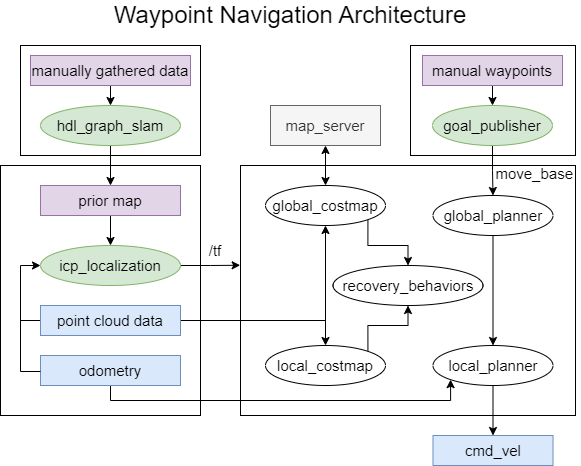}
    \caption{Complete architecture of the proposed waypoint navigation package. Purple boxes represent data that only needs to be handled once. Blue boxes are sent and received by the robot. Green ellipses show the new packages used by this architecture.}
    \label{fig:architecture}
\end{figure}

To enable movement with the localization package, a navigation stack was implemented. The default 2D navigation stack in ROS is the \textit{move\_base}\footnote{http://wiki.ros.org/move\_base} package. The full navigation stack includes obstacle avoidance with 2D costmaps and computes collision-free paths from the current position to a global goal using Dijkstra's algorithm \cite{dijkstra1959note} by default. Once a goal has been set, the navigation stack sends velocity commands to the robot until the goal has been reached. The new \textit{waypoint\_navigation}\footnote{https://github.com/RobustFieldAutonomyLab/waypoint\_navigation} package we have published includes the complete waypoint navigation architecture shown in Fig. \ref{fig:architecture}, with simulated and real-world example launch files using a Jackal ground vehicle with a mounted LiDAR.
Included in our integration of the packages shown in Fig. \ref{fig:architecture} is a waypoint publisher script. The current version of that script accepts 2D goal positions as parameters in a .yaml file, before creating a list of goal positions. When the node is activated, the goals are published in order while waiting for the robot to arrive at its current goal. Once the robot has successfully arrived, the next goal is published until no further goals remain on the list. While simple, this script is effective at supporting repeated autonomous navigation to a series of waypoints.

\section{Experiments}
Multi-session localization for waypoint navigation requires accuracy and precision. Accuracy can be defined by how close the robot is to the target, while precision measures the consistency for a given target.
As we are using a ground vehicle to test our localization algorithm, the waypoints will be defined by their 2D Cartesian coordinates and a quaternion for orientation, $w_i = (x_i, y_i, q_i)$ for the $i$th position of an ordered list $\mathcal{W}_n = (w_1, w_2, \dots , w_n)$ of $n$ waypoints. The Jackal UGV used in our work is described by its pose $s_\tau = (x_\tau, y_\tau, q_\tau)$ at time $t=\tau$. 

\begin{figure*}[ht]
    \centering
    \includegraphics[width=0.925\textwidth]{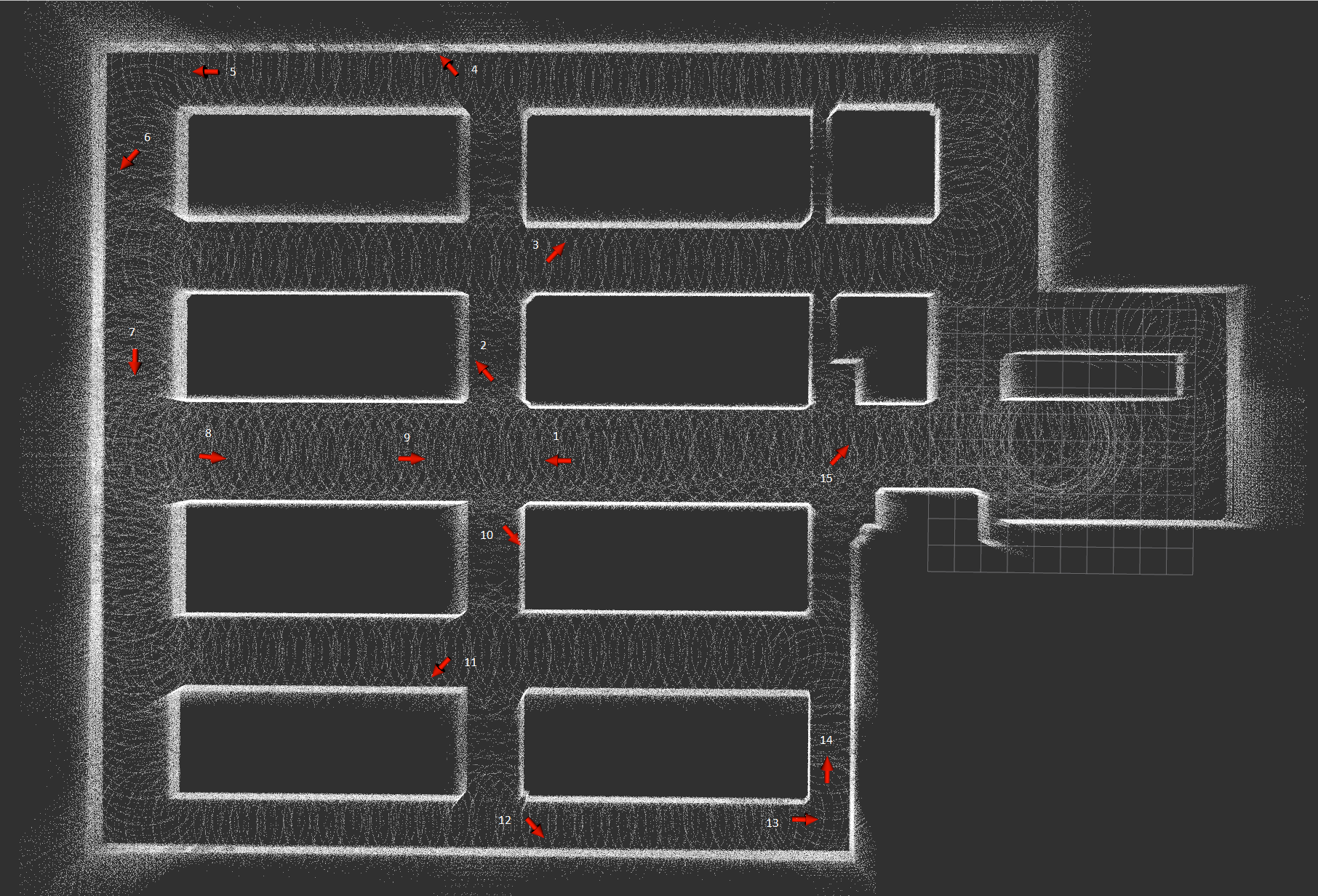}
    \caption{Simulated Gazebo environment with 15 marked waypoints in red. Each cardinal and inter-cardinal direction is represented at least once. The white point cloud is the global reference map used for ICP localization.}
    \label{fig:map_points}
\end{figure*}

\begin{figure*}[ht]
    \centering
    \includegraphics[width=0.8\textwidth]{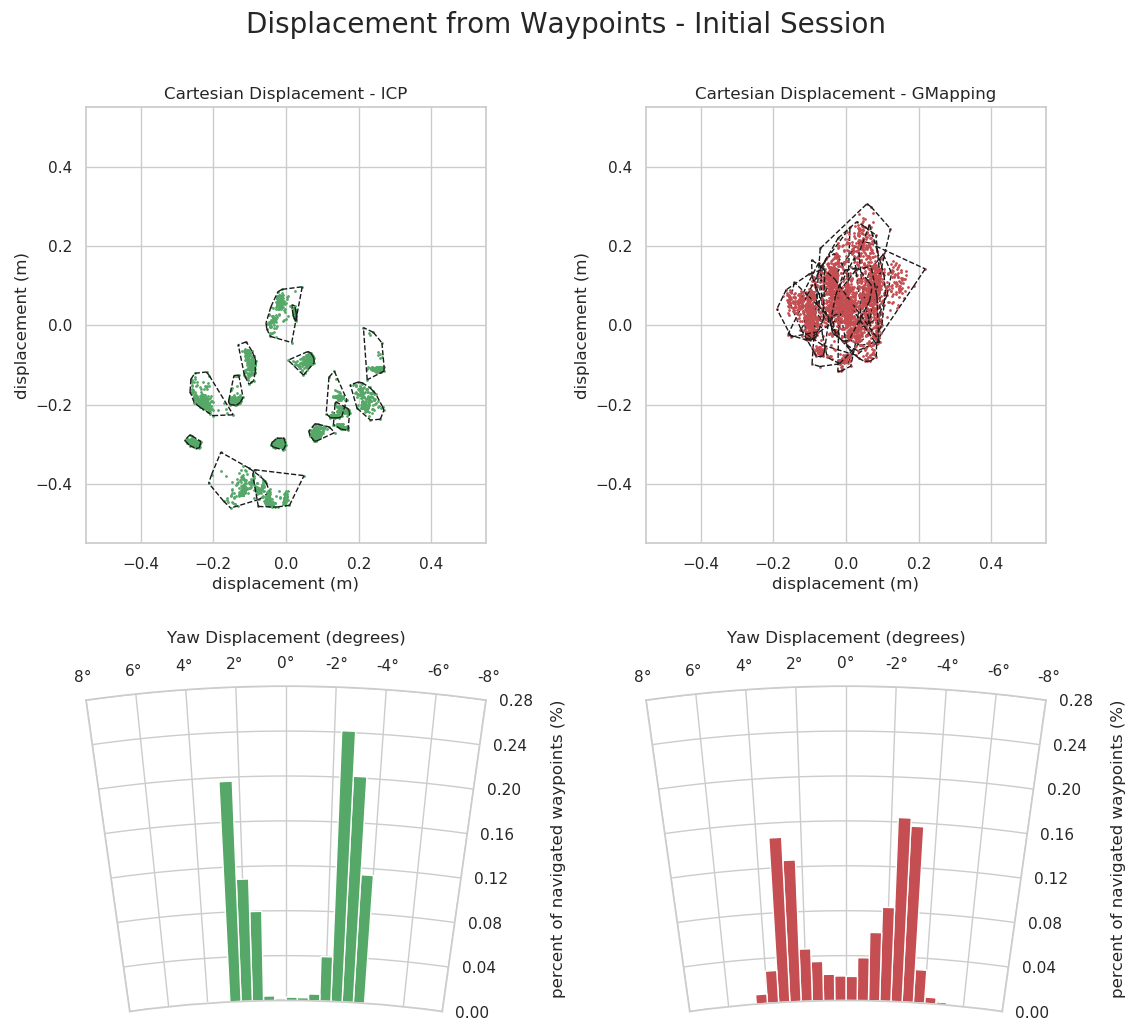}
    \caption{Translational and rotational displacements of Jackal UGV when navigating to waypoints in simulation. Convex hulls around data from each waypoint illustrate navigation precision within the translational displacement plots at top.}
    \label{fig:init}
\end{figure*}

\subsection{Simulation Comparisons}

The proposed waypoint navigation architecture of Fig. \ref{fig:architecture} was built to be simple to use and quick to implement. Given the lack of existing localization algorithm implementations that incorporate navigation, the only practical competitor was SLAM, as implemented to support waypoint navigation in the ROS navigation stack. Many SLAM algorithm implementations exist within ROS, however the simplest to use is GMapping \cite{gmapping}, the default SLAM framework. Therefore, comparisons were made with respect to accuracy and precision of waypoint navigation for both GMapping, and ICP localization (as implemented within our architecture). Simulations were performed in ROS Noetic with an i9-9900K 3.60GHz CPU and 64GB of RAM. ICP trials were given 5 minutes to reach completion, while GMapping required 6 minutes per trial. 

Extensive simulations were performed in Gazebo to compare our proposed use of ICP localization within the architecture of Fig. \ref{fig:architecture} against the conventional usage of  GMapping\footnote{https://wiki.ros.org/gmapping} for waypoint navigation within ROS. While a simulated Jackal UGV manages its state estimation process, the Gazebo simulation environment can provide ground truth. Therefore, to test for accuracy and precision we recorded the ground truth state of the robot when navigation to each designated waypoint was completed. If we denote the time when the robot arrived at waypoint $i$ as $\tau i$, then all our ground truth data for a single trial forms the set $\mathcal{S}_n = \{s_{\tau1}, s_{\tau2}, \dots , s_{\tau n}\}$.

\begin{figure*}[ht]
    \centering
    \includegraphics[width=0.8\textwidth]{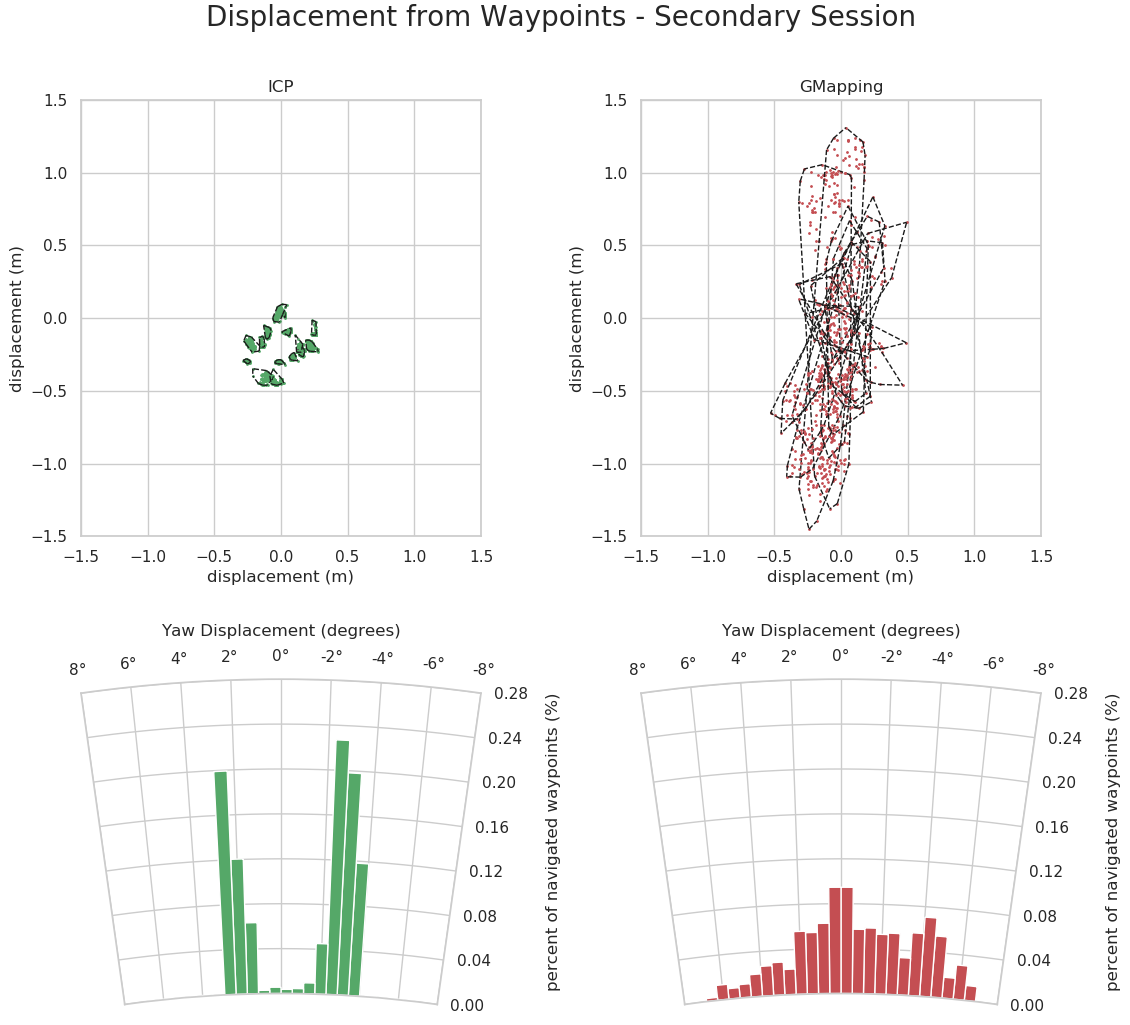}
    \caption{Translational and rotational displacements of Jackal UGV when navigating to waypoints in simulation, during the second navigation session. Convex hulls around data from each waypoint illustrate navigation precision within the translational displacement plots at top.}
    \label{fig:second}
\end{figure*}

A simplified tunnel-like map seen in Fig. \ref{fig:map_points} was used for testing with $n=15$ waypoints chosen to reflect a variety of positions and orientations. Each trial used the same $15$ waypoints assigned in the global reference frame. To ensure a sufficiently large dataset, $m=200$ trials were performed for each waypoint. However, not every trial was successful, resulting in less than $3000$ individual datapoints. If failures occurred in the middle of a trial, data was collected up to the last successfully navigated waypoint. Therefore, we were able to compute a success rate in addition to the accuracy and precision of each framework. 

\begin{algorithm}
    \DontPrintSemicolon
    \KwData{$\mathcal{W}_{15}$}
    $m \gets 200$\;
    $count \gets 0$\;
    $n \gets 15$\;
    \For{$count < m$}{
        begin localization simulation\;
        $i \gets 0$\;
        \For{$i < n$}{
            navigate to $w_i \in \mathcal{W}_{15}$\;
            record robot state\;
            $i++$\;
        }
        end simulation\;
        $count ++$\;
    }
\caption{Simulating 200 trials of navigation to 15 waypoints}
\label{alg:sim}
\end{algorithm}

Multi-session waypoint navigation implies a first session followed by many subsequent sessions. The initial session can be performed by starting a robot in the ideal state, with no errors. However, under autonomous navigation, the robot needs to navigate itself back to the home position and orientation. Any errors accumulated during navigation will affect future sessions if not accounted for. Therefore, we performed two phases of simulation, in which the first phase represents an initial session where the robot was placed exactly at the origin of the map for each trial. By the comparing the final UGV ground truth locations against the assigned waypoints, we created a set $\mathcal{E}_n = \mathcal{S}_n - \mathcal{W}_n = \{e_{1},e_{2},\dots,e_{n}\}$ of target errors. These errors are plotted in Fig. \ref{fig:init} and show how GMapping resulted in more accurate waypoint navigation. However, it is clear that the ICP framework has higher precision, as the convex hulls around each individual waypoint's error $e_i$ form relatively small clusters. 

The second phase of simulation is conducted similarly to the first, with the only change being the starting location of the robot. For these trials, a random error from the first session $e \in \mathcal{E}$ was added to the robot's initial pose. The same map, set of waypoints $\mathcal{W}_n$, and number of trials were performed as in the first round of simulation. However, the results show significantly worse errors for the GMapping framework, while the ICP framework barely had any changes, as seen in Fig. \ref{fig:second}. The success rate of each method was also affected during the change from initial session to second session. For the initial session, ICP had 198 trials reach the first goal point successfully, while only 134 managed to reach the final goal. GMapping had similar values, with 188 reaching the first goal, and 122 completing the entire trial. ICP achieved higher performance in the second round of simulation, with 148 trials reaching the final goal, although only 195 reached the first goal. As expected with the additional error, GMapping had 162 trials reach the first goal, but only 14 trials managed to navigate to all the waypoints. Besides the accuracy and precision outcomes, the robustness of each algorithm demonstrates the benefits of localization using a prior map, for repeated waypoint navigation tasks in indoor environments.

\subsection{Measuring Accuracy}

When navigating to a waypoint, accuracy is a measure of how close the robot was to the true goal. A simple method to compute accuracy is an average of Euclidean distances:

\begin{equation}
    \frac{1}{nm} \sum^{m}_{j=1} \sum^{n}_{i=1} ||z_{ij} - z^*_i||.
    \label{eq:dist}
\end{equation}

For 2D Euclidean distance, $z_{ij} = (x_{ij},y_{ij}) \in s_{\tau i j} \in \mathcal{S}_j$ and the true pose $z^*_i = (x^*_i, y^*_i) \in w_i \in \mathcal{W}$, resulting in an average radial distance. Accuracy was recorded independently of the specific waypoint measured against. 
For our initial session, ICP yielded an accuracy of $0.257994m$, while GMapping has a significantly smaller value at $0.105023m$. Angular orientation accuracy used the same basic equation, with $z_{ij} = q_{ij} \in s_{\tau i j}$ applied to the yaw angle, and once again $z^*_i = q^*_i \in w_i$. As seen in Fig. \ref{fig:init}, both algorithms achieved similar rotational accuracy, which amounted to $0.045672$ radians for ICP and $0.04284$ radians for GMapping. Based on these results, GMapping has higher translational accuracy with comparable rotational accuracy, when compared to the ICP framework, using the exact origin as the robot's starting location for each trial. 


When performing multiple sessions, GMapping was unable to maintain its higher translational accuracy. As seen in Fig. \ref{fig:second}, the ICP framework resulted in nearly the same values as previously, with a translational accuracy of $0.255417m$ and rotational accuracy of $0.045574$ radians. GMapping however became drastically worse at locating waypoints, resulting in a translational accuracy of $0.611964m$. Rotational errors seemed to be more centered with a wider range, leading to an accuracy of $0.043954$ radians.
 

\subsection{Measuring Precision}

Precision can be computed using a similar averaging of the Euclidean norm, with different parameters. This time, errors were divided into subsets based on the waypoint they corresponded to, using Equation \eqref{eq:dist2} for each waypoint $i$:
\begin{equation}
    \frac{1}{m} \sum^{m}_{j=1} ||z_{ij} - z^*_i||.
    \label{eq:dist2}
\end{equation}
Within those subsets, centroids were computed to estimate how tightly packed the errors were. 
For translational precision, the centroids were computed as $z^*_i = (\bar{x}_i,\bar{y}_i)$, while rotational precision was able to directly use the average yaw. Navigation precision was recorded for each waypoint, as seen in Table \ref{tbl:init}. While there is some variability among the results, in the first session, the precision of the ICP framework is superior to GMapping at every waypoint for both translational and rotational precision, with only two angular exceptions of waypoints $5$ and $13$. 

\begin{table}[!b]
\caption{Initial Session Precision}
\centering
\begin{tabular}{||c | c | c | c | c||} 
 \hline
 &\multicolumn{2}{|c |}{2D Norm (m)} & \multicolumn{2}{| c||}{Angular Norm (rad)} \\
 \hline
 Waypoint & ICP & GMap & ICP & GMap \\ [0.5ex] 
 \hline\hline
 1 & \textbf{0.009754} & 0.037302 & \textbf{0.008005} & 0.010106 \\ 
 \hline
 2 & \textbf{0.020213} & 0.043521 & \textbf{0.001827} & 0.014632\\
 \hline
 3 & \textbf{0.012137} & 0.042487 & \textbf{0.002672} & 0.014590\\
 \hline
 4 & \textbf{0.022351} & 0.086543 & \textbf{0.021018} & 0.037307\\
 \hline
 5 & \textbf{0.008230} & 0.059079 & 0.021327 &\textbf{ 0.009029}\\ 
  \hline
 6 & \textbf{0.031140} & 0.089944 & \textbf{0.004168} & 0.035475\\
  \hline
 7 & \textbf{0.026128} & 0.060534 & \textbf{0.005460} & 0.007930\\
  \hline
 8 & \textbf{0.022558} & 0.090067 & \textbf{0.002955} & 0.026483\\
  \hline
 9 & \textbf{0.024736} & 0.049438 & \textbf{0.008284} & 0.036498\\
  \hline
 10 & \textbf{0.010818} & 0.048211 & \textbf{0.002223} & 0.028862\\
  \hline
 11 & \textbf{0.009416} & 0.042898 & \textbf{0.002218} & 0.013759\\
  \hline
 12 & \textbf{0.023257} & 0.047029 & \textbf{0.002852} & 0.013186\\
  \hline
 13 & \textbf{0.011861} & 0.034731 & 0.012552 & \textbf{0.008778}\\
  \hline
 14 & \textbf{0.006096} & 0.035315 & \textbf{0.002980} & 0.030190\\
  \hline
 15 & \textbf{0.030237} & 0.036329 & \textbf{0.002558} & 0.026937\\ [1ex] 
 \hline
\end{tabular}
\label{tbl:init}
\end{table}

\begin{table}[!b]
\caption{Second Session Precision}
\centering
\begin{tabular}{||c | c | c | c | c||} 
 \hline
 &\multicolumn{2}{|c |}{2D Norm (m)} & \multicolumn{2}{| c||}{Angular Norm (rad)} \\
 \hline
 Waypoint & ICP & GMap & ICP & GMap \\ [0.5ex] 
 \hline \hline
 1 & \textbf{0.011429} & 0.721628 & \textbf{0.016104} & 0.032240 \\ 
 \hline
 2 & \textbf{0.018816} & 0.831810 & \textbf{0.001875} & 0.041254\\
 \hline
 3 & \textbf{0.012426} & 0.409511 & \textbf{0.002716} & 0.037715\\
 \hline
 4 & \textbf{0.022374} & 0.453125 & \textbf{0.024044} & 0.045671\\
 \hline
 5 & \textbf{0.008631} & 0.466506 & \textbf{0.021348} & 0.024298\\ 
  \hline
 6 & \textbf{0.027262} & 0.520502 & \textbf{0.005759} & 0.035139\\
  \hline
 7 & \textbf{0.025202} & 0.469325 & \textbf{0.007286} & 0.016256\\
  \hline
 8 & \textbf{0.025000} & 0.426370 & \textbf{0.005387} & 0.021191\\
  \hline
 9 & \textbf{0.026525} & 0.332063 & \textbf{0.006611} & 0.030635\\
  \hline
 10 & \textbf{0.010797} & 0.287184 & \textbf{0.002530} & 0.035593\\
  \hline
 11 & \textbf{0.010091} & 0.295293 & \textbf{0.002703} & 0.021767\\
  \hline
 12 & \textbf{0.024989} & 0.291501 & \textbf{0.002888} & 0.031338\\
  \hline
 13 & \textbf{0.012936} & 0.191098 & \textbf{0.015607} & 0.021560\\
  \hline
 14 & \textbf{0.005804} & 0.201783 & \textbf{0.002776} & 0.019334\\
  \hline
 15 & \textbf{0.031349} & 0.119085 & \textbf{0.001721} & 0.025049\\ [1ex] 
 \hline
\end{tabular}
\label{tbl:second}
\end{table}

Similar to the decline in accuracy, GMapping suffered drastically with respect to the precision realized in the translational errors of its second session. Most waypoints saw order-of-magnitude worse results, while all angular precision measurements were worse than those from the ICP algorithm, as seen in Table \ref{tbl:second}.

 \begin{figure*}[t]
    \centering
    \includegraphics[width=0.99\textwidth]{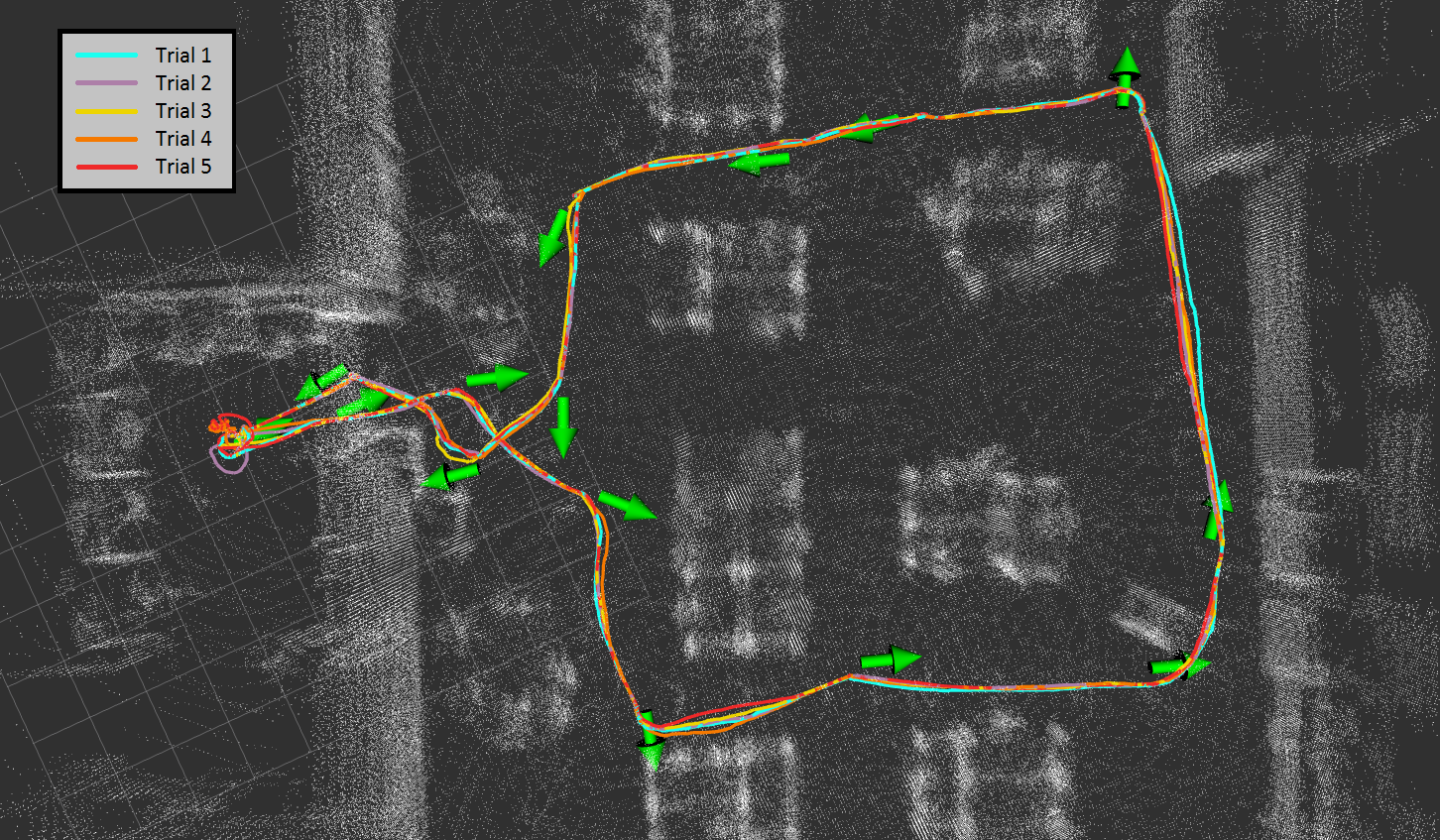}
    \caption{Five consecutive execution traces of autonomous waypoint navigation on the Jackal UGV using the proposed architecture of Fig. \ref{fig:architecture} in Stevens' ABS Engineering Center. Fifteen waypoints marked with green arrows were sent via script to run autonomously. Each new trial was started by resetting the localization software, but the hardware was not moved, to ensure true multi-session navigation. A video of this process can be viewed here:\\ \url{https://youtu.be/vSRrwgDN9kg}}
    \label{fig:abs_arrows}
\end{figure*}

\subsection{Real World Experiments}

\begin{figure}[!b]
    \centering
    \includegraphics[width=0.45\textwidth]{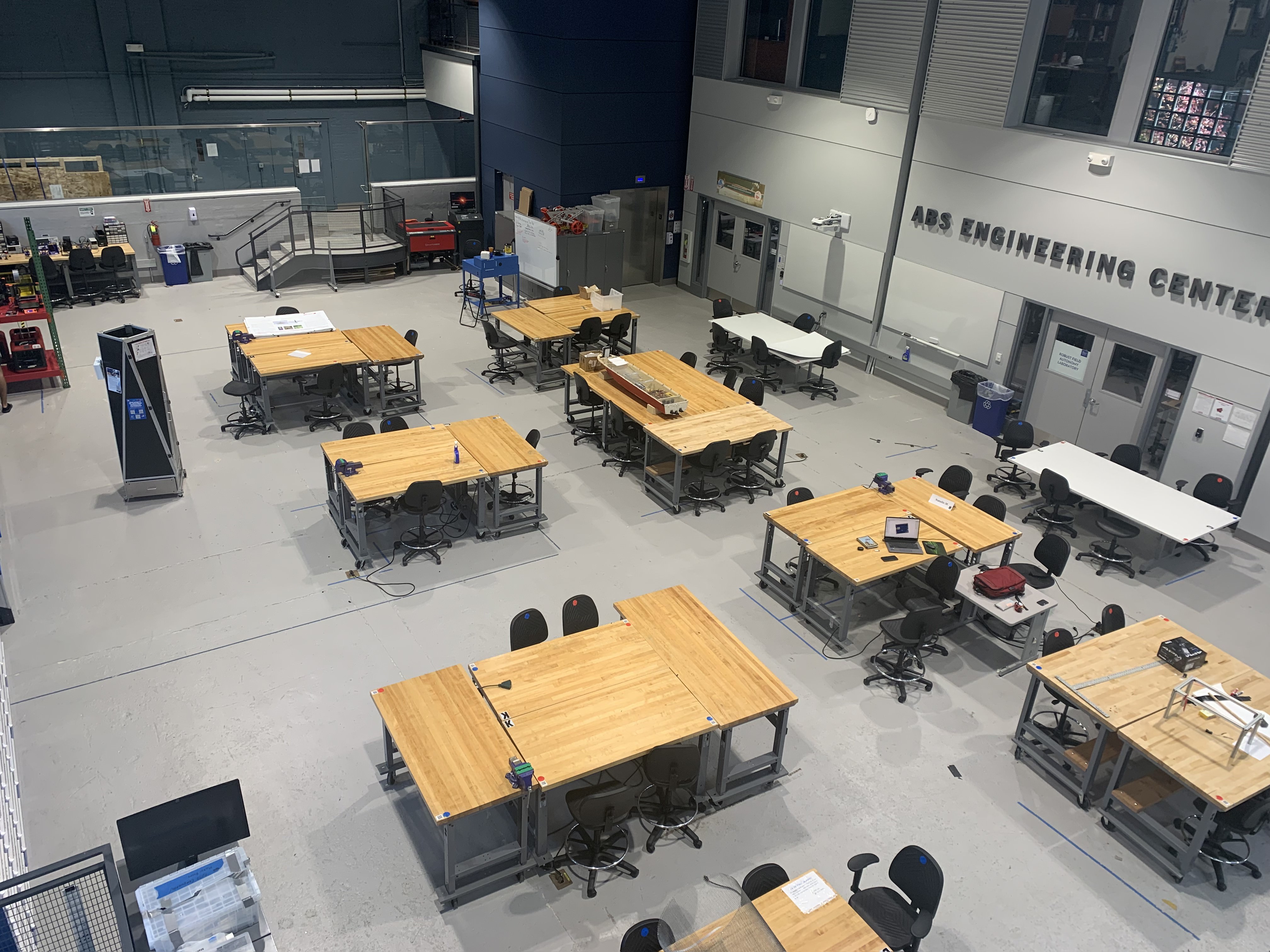}
    \caption{ABS Engineering Center used for real world testing of ICP localization framework on Jackal UGV.}
    \label{fig:abs}
\end{figure}

ICP localization was implemented on a Jackal UGV using a global reference map of the ABS Engineering Center at Stevens Institute of Technology seen in Fig. \ref{fig:abs}. A script published the same waypoints for five consecutive trials. The final waypoint was placed at the origin of the map so that the robot could return autonomously to its start location. To simulate multi-session navigation, the localization algorithm was restarted remotely before each trial began, however the hardware was not touched between trials. All five trials can be seen in Fig. \ref{fig:abs_arrows} where the pathways varied, but the waypoints were reached precisely.

All of the preparations to perform real world testing, and its execution, occurred on the same day. LiDAR data was gathered by manually driving the Jackal UGV around the ABS Engineering Center while recording. For complete coverage, the manual path taken was more comprehensive than our final autonomous path. The environment was small enough that our offline SLAM solution produced an accurate prior map to use in less than 10 minutes (shown in white in Fig. \ref{fig:abs_arrows}). After updating the localization parameters, waypoint navigation testing began. Waypoints were  manually placed on the global reference map via rviz. If the waypoints were reasonable (e.g., appeared to be collision-free), they were added to the autonomous navigation script. Once the script was complete, everything was put in place for autonomous navigation around the indoor environment. The total process from gathering the prior map data to defining waypoints and autonomously navigating to them took less than 4 hours.

The reference map used in this experiment included some extra data of dynamic obstacles, such as team members following the robot around. During each navigation trial, students were actively moving around this workspace as well, resulting in lidar scans that were not perfectly matched to
the prior map. However, ICP localization combined with odometry achieved enough point to point matches to overcome any mismatches due to moving obstacles, which resulted in accurate localization even with these discrepancies. Obstacle avoidance was restarted at the beginning of each trial, and only current data was used to define obstacles, which enabled the Jackal to avoid new objects that did not exist when the global reference map was created.

\section{Conclusion}
Using publicly available software packages, our team was able to produce a reliable and quick system for building a global reference map and performing localization that was sufficient for autonomous waypoint navigation in a previously unknown environment. The global reference map does not need to be exactly the same as the live data, thanks to the robustness of ICP combined with odometry data. This is particularly helpful in environments with constantly changing floor spaces and people actively walking around. While we were able to tackle localization for ground vehicles, the default move\textunderscore base navigation package had limited success at avoiding obstacles across multiple sessions, under the accumulation of errors. We hope that the proposed architecture can serve as a foundational capability upon which the robotics community can achieve more complex task execution in GPS-denied indoor environments. 


%

%
%
\bibliographystyle{ieeetr}
\bibliography{references}
\newpage

\end{document}